\documentclass[10pt,twocolumn,letterpaper]{article}

\usepackage{cvpr}   
\definecolor{cvprblue}{rgb}{0.21,0.49,0.74}
\usepackage[pagebackref,breaklinks,colorlinks,allcolors=cvprblue]{hyperref}

\def\paperID{*****} % *** Enter the Paper ID here
\def\confName{CVPR}
\def\confYear{2026}

\usepackage[most]{tcolorbox}

\tcbset{
  titlebar/.style={
    colback=gray!35,
    colframe=gray!35,
    boxrule=0pt,
    left=8pt,right=8pt,top=6pt,bottom=6pt,
    sharp corners,
  },
}

\tcbset{
  contentbox/.style={
    colback=gray!10,
    colframe=black!50,
    arc=2mm,
    boxrule=0.6pt,
    left=8pt,right=8pt,top=8pt,bottom=8pt,
  },
}

\usepackage{float}   % 放在导言区

\title{\raisebox{-1.5ex}{\includegraphics[height=1.8em]{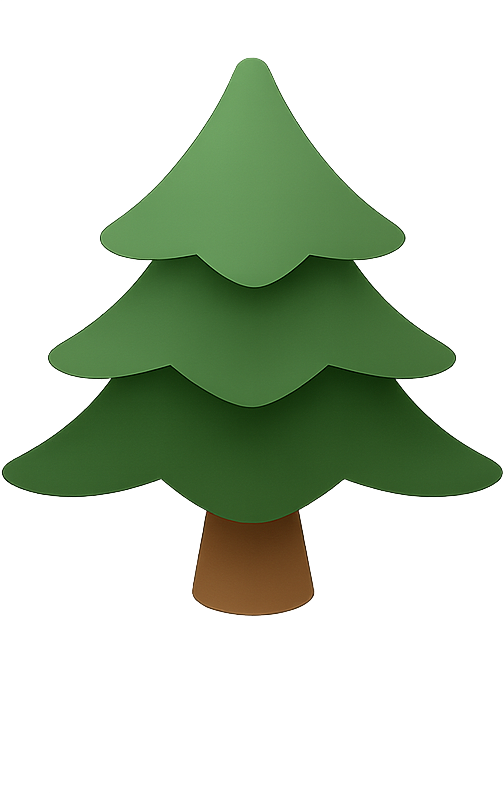}}\hspace{0.2em}\normalfont\bfseries AgriGPT-VL: An Unified Suite for Agricultural Vision–Language Understanding}

\author{
Bo Yang\\
Zhejiang University\\
{boyang30@zju.edu.cn}
\and
Lanfei Feng\\
Zhejiang University\\
{22451116@zju.edu.cn}
\and
Yunkui Chen\\
Zhejiang University\\
{22351048@zju.edu.cn}
\and
Yu Zhang\\
Zhejiang University\\
{22421173@zju.edu.cn}
\and
Xiao Xu\\
Zhejiang University\\
{3200105334@zju.edu.cn}
\and
Jianyu Zhang\\
Zhejiang University\\
{jianyu.zhang@zju.edu.cn}
\and
Nueraili Aierken\\
Zhejiang University\\
{nureli@zju.edu.cn}
\and
Runhe Huang\\
Hosei University\\
{rhuang@hosei.ac.jp}
\and
Hongjian Lin\\
Zhejiang University\\
{linhongjian@zju.edu.cn}
\and
Yibin Ying\\
Zhejiang University\\
{ibeying@zju.edu.cn}
\and
Shijian Li\textsuperscript{*}\\
Zhejiang University\\
{shijianli@zju.edu.cn}
}

\begin{document}
\maketitle
\begin{abstract}
Despite rapid advances in multimodal large language models, agricultural applications remain constrained by the scarcity of domain-tailored models, curated vision–language corpora, and rigorous evaluation. To address these challenges, we present the \textbf{AgriGPT-VL Suite}, a unified multimodal framework for agriculture. Our contributions are threefold. First, we introduce \textbf{Agri-3M-VL}, the largest vision–language corpus for agriculture to our knowledge, curated by a scalable multi-agent data generator; it comprises 1M image–caption pairs, 2M VQA (Visual Question Answering) pairs, 50K expert-level VQA, and 15K GRPO reinforcement learning dataset. Second, we develop \textbf{AgriGPT-VL}, an agriculture-specialized vision–language model trained via a progressive curriculum of textual grounding, multimodal shallow/deep alignment, and GRPO refinement. This method achieves strong multimodal reasoning while preserving text-only capability. Third, we establish \textbf{AgriBench-VL-4K}, a compact yet challenging evaluation suite with a multi-metric evaluation and an LLM-as-a-judge framework.  Experiments show that AgriGPT-VL outperforms leading general-purpose VLMs on AgriBench-VL-4K, achieving higher pairwise win rates in the LLM-as-a-judge evaluation. Meanwhile, it remains competitive on the text-only AgriBench-13K with no noticeable degradation of language ability. Ablation studies further confirm consistent gains from our alignment and GRPO refinement stages. 
\end{abstract}

\begin{figure}[t]
\centering
\begin{minipage}{1\linewidth}
    \centering
    \includegraphics[width=\linewidth]{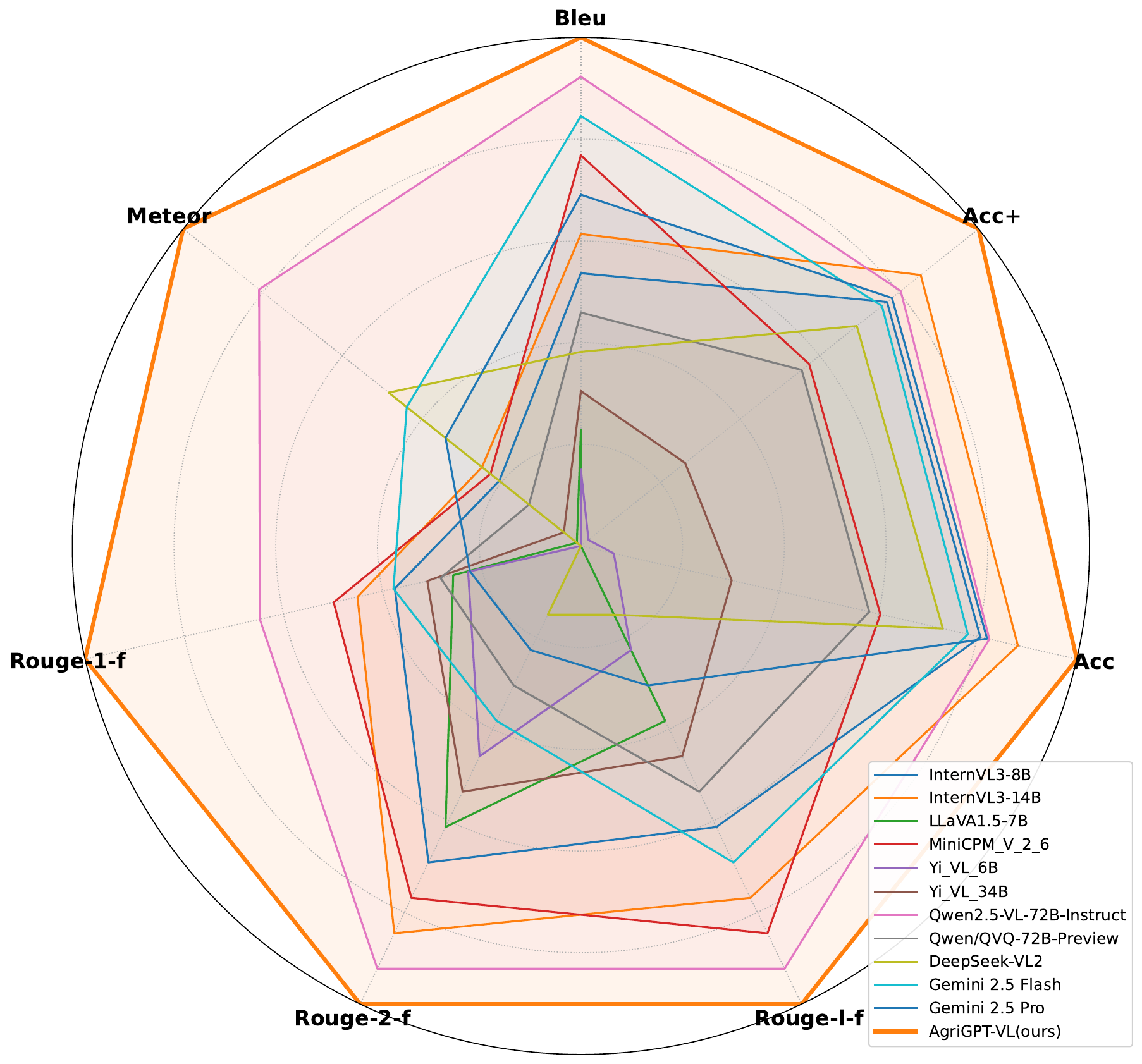}
    \caption{AgriGPT-VL achieves leading performance on AgriBench-VL-4K.}
    \label{fig:radar}
\end{minipage}
\end{figure}

\begin{figure*}[t]
\centering
\includegraphics[width=0.95\linewidth]{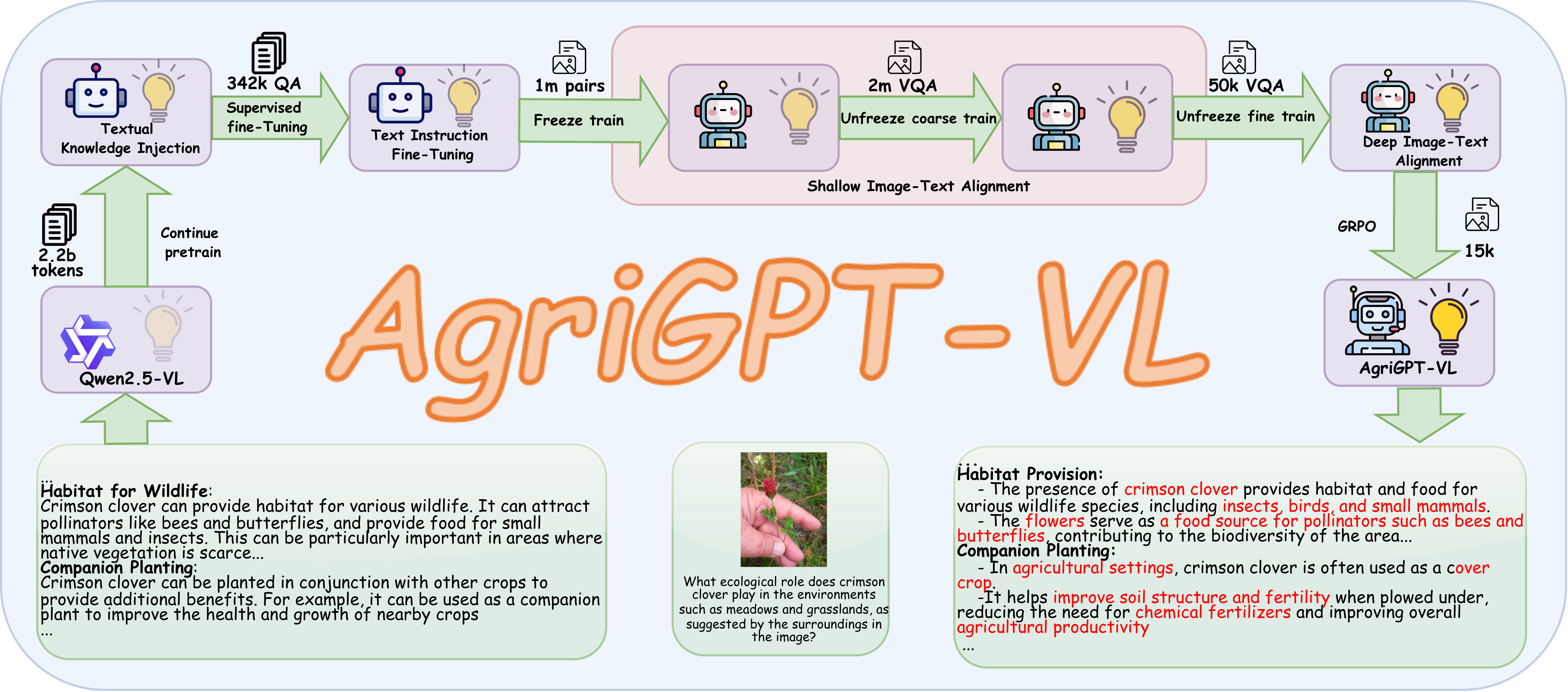}
% \caption{Overview of the AgriGPT-VL training pipeline and curriculum-based model evolution.}
% \caption{Overview of the AgriGPT-VL training pipeline, showing the progression from textual tuning to shallow and deep image–text alignment, followed by GRPO refinement.}
\caption{Overview of the AgriGPT-VL training pipeline, showing the progression from textual tuning to shallow and deep image–text alignment, followed by GRPO refinement, together with the datasets used at each stage.
}
\label{fig:overview}
\end{figure*}

\section{Introduction}
\label{sec:intro}

The convergence of AI with critical sectors like agriculture presents a significant opportunity to address global challenges such as food security and sustainable resource management \citep{Swaminathan2001,Foley2011,Clapp2020}. With the increasing challenges posed by climate change, resource scarcity, and population growth, intelligent agricultural decision-making is becoming indispensable \citep{Godfray2010,Rockstrom2017,FanRue2020}. In recent years, multimodal large language models (MLLMs) have demonstrated remarkable progress in integrating vision and language, enabling tasks such as captioning, visual question answering (VQA), and multimodal reasoning \citep{Yin2023MLLMsurvey,Achiam2023GPT4,Chen2024LVLMeval}. While Multimodal Large Language Models (MLLMs) excel at integrating vision and language on general web data \citep{Schuhmann2022LAION5B,Du2022VLPsurvey}, they are ill-equipped for the agricultural domain. The knowledge required for tasks in crop and soil science is highly specialized and absent from standard pre-training corpora \citep{Kamilaris2018DLagri,Wolfert2017BigData}. Consequently, existing MLLMs struggle with agricultural terminology, exhibit factual inaccuracies, and fail to provide reliable, context-aware support for real-world farming  operations \citep{wu2024newagronomists,Rezayi2022AgriBERT,Yang2025AgriGPT}.

Several attempts have been made to build agricultural language models, such as AgriBERT \citep{Rezayi2022AgriBERT}, AgriLLM \citep{Didwania2024AgriLLM}, AgroLLM \citep{Samuel2025AgroLLM}, and AgroGPT \citep{Awais2025AgroGPT}. These efforts show the value of domain-specific adaptation but are often constrained to text-only settings or narrow task coverage. AgriGPT \citep{Yang2025AgriGPT}, introduced the first agriculture-specialized LLM ecosystem with a curated instruction dataset(Agri-342K), a retrieval-enhanced reasoning module (Tri-RAG), and a benchmark suite (AgriBench-13K). While effective for textual reasoning, AgriGPT lacked visual reasoning ability and thus could not address multimodal agricultural tasks such as pest recognition or crop diagnosis. On the other hand, general-purpose MLLMs such as InternVL \citep{Chen2024InternVL}, Qwen-VL \citep{QwenTeam2024QwenVL}, Gemini \citep{Achiam2023GPT4}, and LLaVA \citep{Liu2023LLaVA} demonstrate strong vision--language capabilities but are trained primarily on internet-scale data describing common objects, scenes, and events, which fail to capture agricultural semantics. As a result, these models suffer from hallucinations, poor transferability, and lack of reasoning ability in agriculture-specific scenarios. Related domains such as medicine developed specialized multimodal LLMs, highlighting the need for a comparable ecosystem in agriculture.

\noindent\textbf{Our contributions can be summarized as follows:}
\begin{itemize}
\item \textbf{Agri-3M-VL Dataset \& Data Generator.}
We build a transferable, reusable multi-agent Data Generator and use it to curate \textbf{Agri-3M-VL}: \textbf{1M} image--caption pairs, \textbf{2M} high-quality VQA pairs, \textbf{50K} expert-level VQA, and \textbf{15k} rewarded GRPO reinforcement learning dataset.

To the best of our knowledge, this is the largest agriculture vision--language corpus to date.

\item \textbf{AgriGPT-VL \& Curriculum Training.}
Using a progressive curriculum, we train the agriculture-specialized VL model \textbf{AgriGPT-VL}, as shown in figure~\ref{fig:radar}, which surpasses most flagship models in capability.

\item \textbf{AgriBench-VL-4K \& Evaluation Framework.}
We construct a comprehensive and challenging benchmark with \textbf{2{,}018} open-ended VQA and \textbf{1{,}858} image-grounded single-choice questions (two per image for cross-consistency).

\item \textbf{Open Resources. }All resources will be released as open-source to ensure full reproducibility and enable deployment in low-resource agricultural scenarios.

\end{itemize}

\begin{figure*}[t]
\centering
\includegraphics[width=0.95\linewidth]{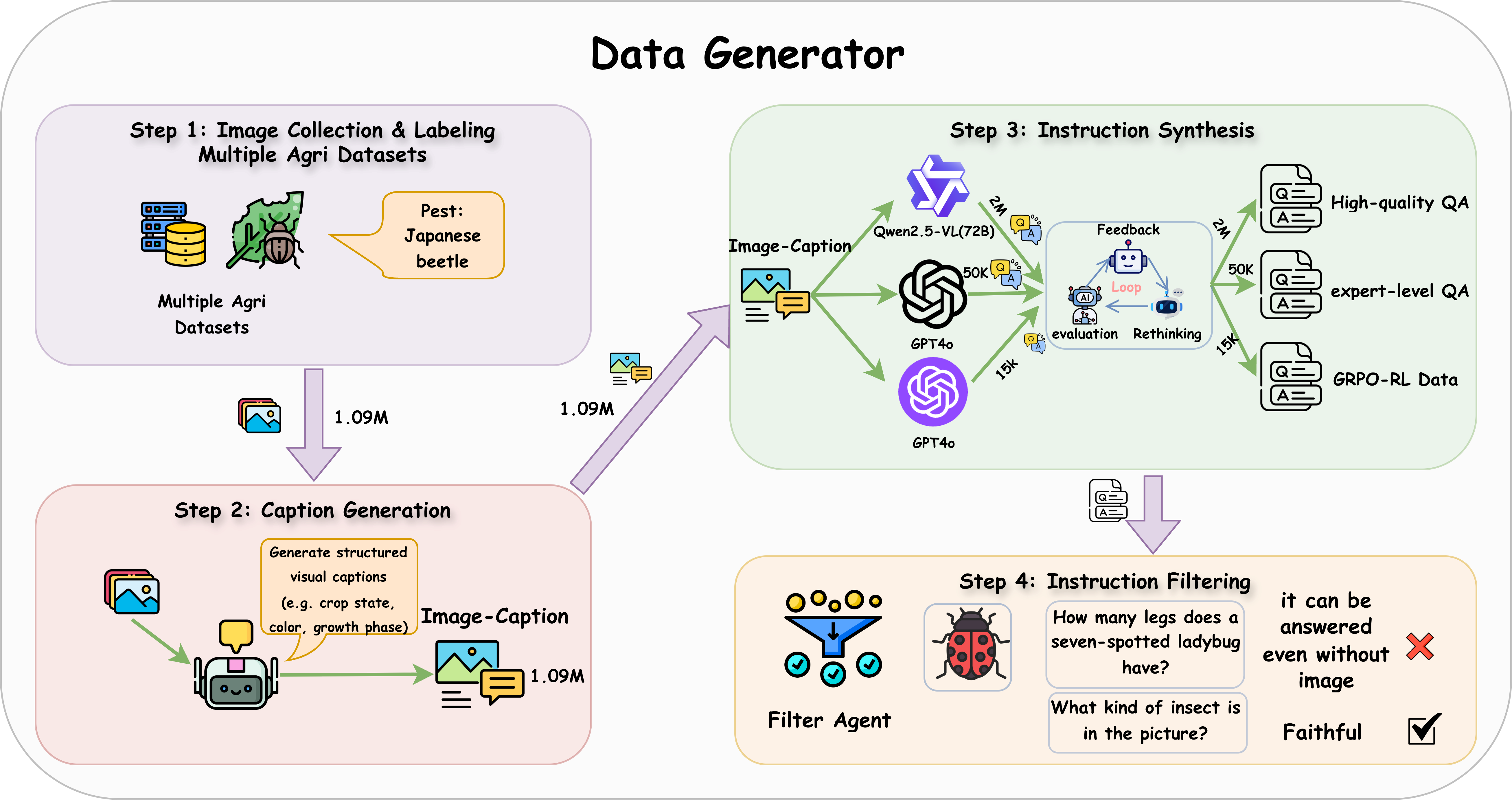}
\caption{Data Generator: A multimodal instruction data generation pipeline.}
\label{fig:data generator}
\end{figure*}

\section{Related Work}
\label{sec:formatting}

\subsection{Text-Only Language Models in Agriculture}
Pioneering work in agricultural AI largely focused on the language modality. Early models such as AgriBERT \citep{Rezayi2022AgriBERT} adapted language model pre-training to domain-specific text corpora. Subsequent efforts, including AgriLLM \citep{Didwania2024AgriLLM}, AgroLLM \citep{Samuel2025AgroLLM}, and AgriGPT \citep{Yang2025AgriGPT}, advanced this paradigm by developing large-scale instruction datasets like Agri-342K and text-only benchmarks such as AgriBench-13K \citep{Yang2025AgriGPT}. For instance, Zhu et al.~\citep{zhu2024harnessing} reviewed the progression of text-only and multimodal agricultural LLMs, highlighting the transition from domain adaptation to instruction-based fine-tuning. Moreover, Yu and Lin~\citep{yu2024framework} proposed a framework leveraging LLMs for agricultural knowledge inference and consultation, suggesting broader utility beyond QA. While these models demonstrated strong textual understanding, their primary limitation was the absence of visual reasoning ability, restricting their applicability to tasks that do not require visual interpretation.

\subsection{Emergence of Multimodal Agricultural Systems}

The integration of visual data marked a critical evolution in agricultural AI. Foundational datasets like PlantVillage \citep{Hughes2015PlantVillage} and IP102 \citep{wu2019ip102} provided large-scale image collections for specific recognition tasks, such as pest and disease identification. More recent works have begun to build multimodal models and benchmarks with broader capabilities. For instance, Agri-LLaVA \citep{wang2024agri}, AgriCLIP \citep{nawaz2024agriclip}, and LLMI-CDP \citep{wang2025large} introduced vision--language abilities, while datasets like VL-PAW \citep{Yu2025VL-PAW} and benchmarks like AgMMU \citep{gauba2025agmmu}, AgroBench \citep{shinoda2025agrobench}, and AgriEval \citep{yan2025agrieval} introduced tasks such as VQA and captioning. Other studies such as \citet{zhu2024harnessing} provide a systematic review of the current landscape, while \citet{yu2024framework} and \citet{arshad2025leveraging} explore concrete frameworks or empirical evaluations of VLMs in agricultural use cases. However, these multimodal resources often remain limited in scale, are restricted to narrow recognition tasks, or lack rigorous, large-scale quality control, representing disparate efforts rather than a cohesive foundation.

\subsection{The Need for a Unified Vision--Language Ecosystem}
The limitations of prior work highlight a clear need for a comprehensive and unified framework. While previous efforts have made valuable contributions to datasets, models, or benchmarks individually, progress has been hampered by the lack of a single ecosystem that integrates all three components at scale. To address this fragmentation, our work introduces a cohesive suite of resources. Our \textbf{Agri-3M-VL dataset} provides scale and quality; our \textbf{AgriGPT-VL} model handles complex reasoning beyond simple recognition; and our \textbf{AgriBench-VL-4K} benchmark enables robust, multifaceted evaluation. Together, these components form the kind of unified foundation we argue is necessary for the next generation of agricultural AI.

\section{AgriGPT-VL}
\label{sec:formatting}

\subsection{Agri-3M-VL Dataset}

Constructing training data is a fundamental challenge in developing multimodal large language models. To address this, we introduce the Data Generator, a transferable paradigm for systematically transforming raw images into high-quality multimodal instructions. The generator is designed not only for agriculture but also as a generalizable methodology that can be applied to other scientific domains where multimodal resources remain scarce or noisy.

\begin{figure}[t]
\centering
\begin{minipage}{1\linewidth}
    \centering
    \includegraphics[width=\linewidth]{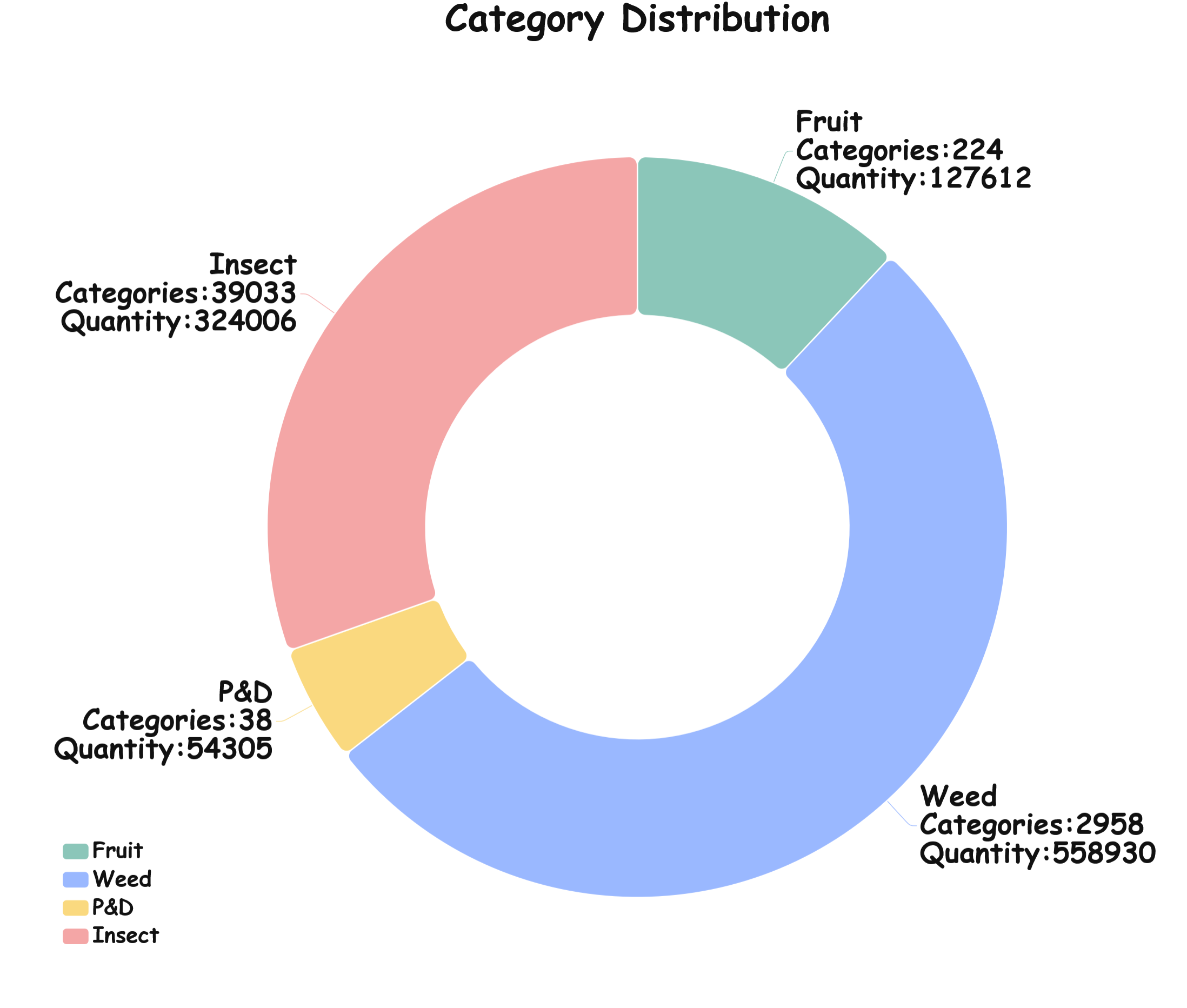}
    \caption{Category distribution of the dataset.}
    \label{fig:dataset distribution}
\end{minipage}
\end{figure}

As shown in Figure~\ref{fig:dataset distribution}, we aggregated a wide range of datasets covering pests and diseases, insects, crops, weeds, and fruits. Specifically, the PlantVillage dataset contains 54,305 images across 38 classes~\citep{plantvillage2019}. For insect-related data, we included 6,878 images covering 166 fine-grained insect species from the Species196 dataset~\citep{he2023species196}, and the Insect Foundation dataset with 317,128 images spanning 38,867 fine-grained insect classes~\citep{insectfoundation2020}, totaling 324,006 images and 39,033 classes. In the crop and weed domain, the SelectDataset provides 558,930 images over 2,958 categories~\citep{selectdataset2021}. For fruits, we incorporated Fruits-360 with 97,255 images and 206 categories~\citep{fruits3602018}, and Fresh-Rotten Fruit with 30,357 images and 18 categories~\citep{freshrotten2022}, amounting to a combined 157,969 images and 224 classes. Altogether, these datasets cover \textbf{1,064,853 } images and \textbf{42,253 }  fine-grained categories, nearly encompassing the full agricultural visual landscape.

However, these raw datasets suffer from several limitations: many lack descriptive annotations, exhibit inconsistent labeling, and cannot be directly used for multimodal model training. These shortcomings necessitate our proposed Data Generator, which systematically transforms such raw images into structured, instruction-ready corpora. Through several stages of processing, the Data Generator enables the creation of a large-scale, high-quality multimodal training corpus suitable for agricultural vision--language modeling.

\begin{figure*}[t]
\centering
\includegraphics[width=0.95\linewidth]{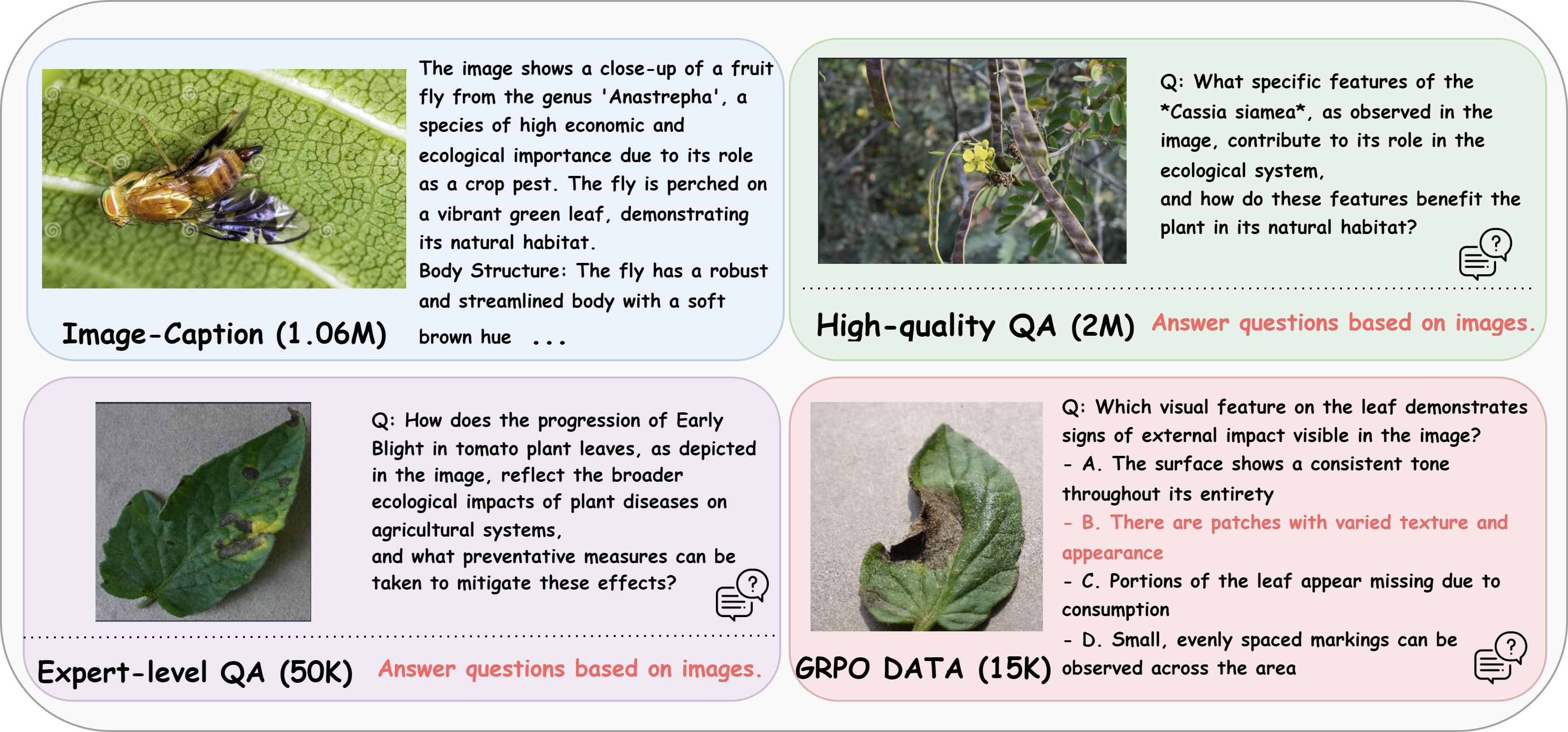}
\caption{The four types of hierarchical training data constructed for AgriGPT-VL.}
\label{fig:shuju caseshow}
\end{figure*}

As shown in figure~\ref{fig:data generator}, the Data Generator transforms multi-source agricultural images into instruction-ready corpora via four stages—caption generation, instruction synthesis, multi-agent refinement and instruction filtering yielding ~1M image captions, 2M high-quality VQA, a 50K expert-level VQA, and 15k GRPO reinforcement learning dataset.The detailed high-quality VQA are illustrated in figure~\ref{fig:shuju caseshow}.

\noindent\textbf{(1) Caption Generation.} For the collected images spanning pests and diseases, insects, weeds, and fruits, we first generate structured visual captions. These captions describe observable attributes such as crop growth stage, leaf color, fruit maturity, or pest morphology. For example, an image of diseased tomato leaves is captioned with information about lesion color and spread, while a fruit image records ripeness stage and external texture. In total, this stage yields about 1 million image–caption pairs, providing a descriptive foundation for subsequent instruction synthesis. The significance of this step is that captions transform raw visual data into semantically rich text, enabling downstream models to link domain-specific imagery with meaningful language.

\noindent\textbf{(2) Instruction Synthesis.} Building upon the image–caption pairs, we employ large vision–language models (e.g., Qwen2.5-VL 72B, GPT-4o) to generate diverse instructions and answers. This stage produces multiple types of VQA: high-quality factual queries, expert-level reasoning tasks, and interactive multimodal dialogues. For instance, a weed image may lead to questions such as “What species of weed is shown?” (recognition) or “What is the likely impact of this weed on crop yield?” (reasoning). Altogether, we synthesize approximately 2 million VQA samples, covering both open-ended and single-choice formats. This step is essential because it elevates the dataset from simple recognition to instruction-following reasoning, directly aligning with the needs of multimodal LLMs.

(3) \textbf{Multi-Agent Refinement.} To further construct high-quality VQA data on top of the Image--Caption corpus, we design a protocol-guided multi-agent refinement architecture and adopt Qwen2.5-72B as the core execution model to balance efficiency and computational cost. The architecture consists of three expert agents---\textbf{Feedback}, \textbf{Evaluation}, and \textbf{Rethinking}---which collaborate through a structured ``generate $\rightarrow$ assess $\rightarrow$ revise'' loop. The \textbf{Feedback agent} generates an initial question--answer draft based on the image and its caption, with protocol constraints requiring that the content reference only visible entities and attributes, avoid out-of-image inferences, and maintain clear visual evidence. The \textbf{Evaluation agent} agent performs structured quality assessment along correctness, clarity, and completeness (each scored within $[0,1]$) and also produces concise diagnostic feedback identifying inconsistencies, ambiguity, or insufficient grounding; any score below the unified threshold of $0.85$ triggers revision. The \textbf{Rethinking agent} revises the question and answer according to this feedback and performs a self-consistency check by independently generating the answer three times, requiring an agreement of at least $0.85$; otherwise, further revision is initiated. This protocolized refinement loop typically converges in approximately $2.3$ rounds, and we retain only samples that satisfy all quality requirements, yielding about $2$M high-quality VQA pairs. We further select $50$K samples for multi-round verification and polishing using GPT-4o to construct a high-quality supervised fine-tuning subset, and additionally create a $15$K GRPO preference dataset, also with GPT-4o, for reinforcement learning. Empirically, about $8\%$ of initial drafts are filtered due to insufficient correctness or grounding, and manual inspection of a $5$K subset confirms that all three quality dimensions consistently exceed $0.86$, demonstrating the effectiveness of the multi-agent refinement architecture in reducing hallucination and improving data reliability.

\noindent\textbf{(4) Instruction Filtering.} Finally, we introduce a \textbf{Filter agent} to automatically identify and discard instructions that are irrelevant to the image or potentially hallucinated. The agent evaluates each instruction along three protocol-defined dimensions—correctness, image-dependence, and grounding validity—and marks a sample for removal whenever any dimension falls below the required threshold. As a result, generic questions unrelated to the image (e.g., ``How many legs does a seven-spotted ladybug have?'') are correctly filtered out, while questions that genuinely rely on visual evidence (e.g., ``What kind of insect is in the picture?'') are retained. Quantitatively, about $0.6\%$ of the initial instructions are accurately identified as irrelevant or factually inconsistent; in a human audit of $300$ randomly sampled instructions, none of the filtered data exhibited such issues. This filtering step further strengthens factual alignment, suppresses hallucination propagation, and improves the overall trustworthiness of the training data.

 Each stage is complementary: caption generation provides semantic grounding, instruction synthesis injects reasoning diversity, multi-agent refinement structures feedback-driven selection, and instruction filtering enforces factual reliability. Together, they form a robust agricultural multimodal dataset that not only supports AgriGPT-VL training but also serves as a blueprint for dataset construction in other scientific domains.

\begin{figure}[t]
\centering
\begin{minipage}{1\linewidth}
    \centering
    \includegraphics[width=\linewidth]{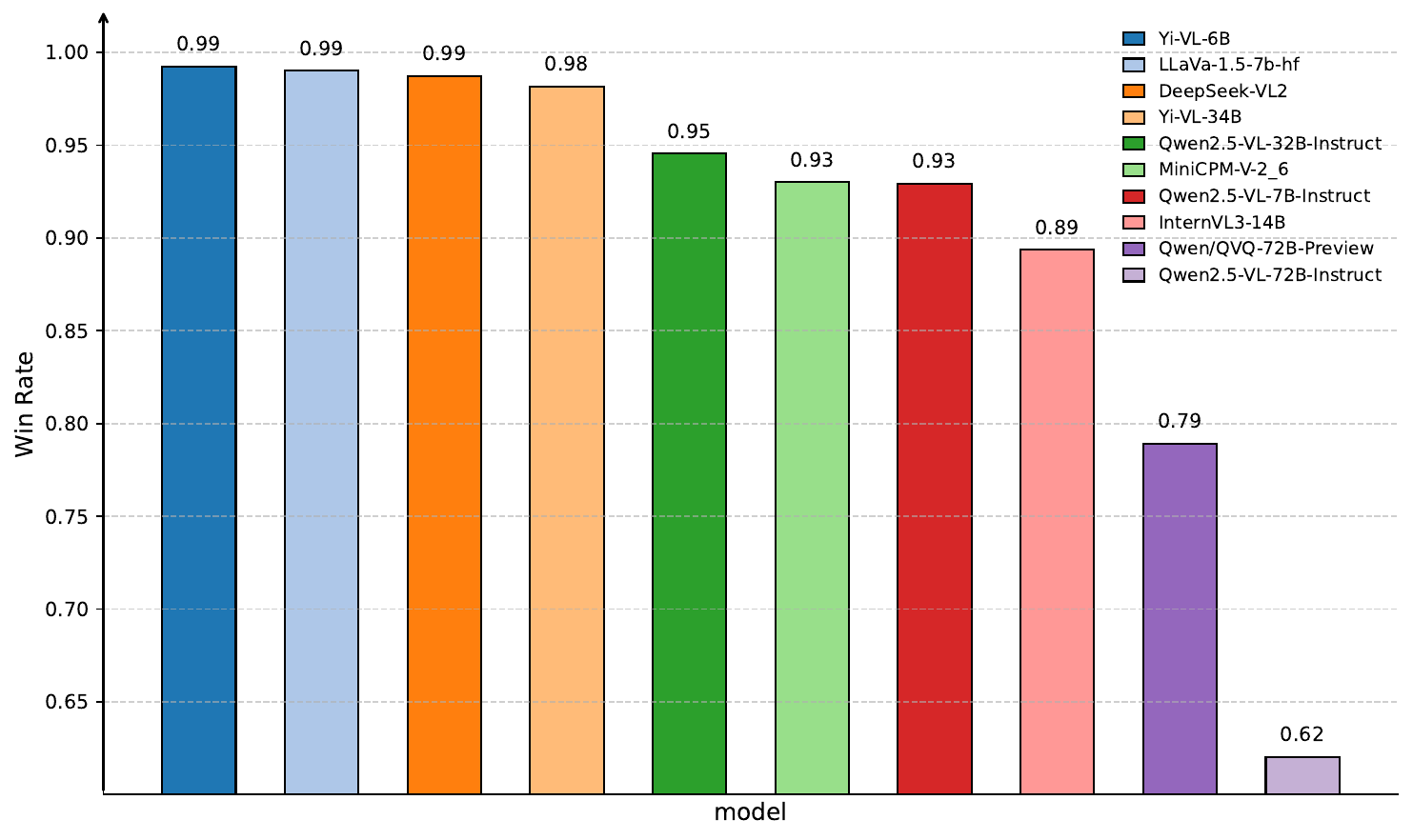}
        \caption{Pairwise win rate of vision-language models vs. AgriGPT-VL (Qwen2.5-VL-72B-as-a-judger)}
    \label{fig:LLM based Judger}
\end{minipage}
\end{figure}

\subsection{AgriGPT-VL Model Training}

As shown in figure~\ref{fig:overview}. This section details our training paradigm for AgriGPT-VL, which follows a progressive curriculum: textual grounding first, then vision–language alignment. We first consolidate domain knowledge and instruction style on text-only data, and then align vision and language on synthesized multimodal supervision with an easy-to-hard schedule.

\textbf{Stage-1 (Text-only).} Starting from Qwen2.5-VL, we conduct continual pretraining on about 200K documents ($\approx$ 2.2B tokens) to inject agricultural terminology and background knowledge, followed by supervised instruction tuning on Agri-342K \citep{Yang2025AgriGPT}. A held-out split of AgriBench-13K \citep{Yang2025AgriGPT} is used for early stopping and calibration prior to multimodal alignment.

\textbf{Stage-2 (Curricular Alignment on Synthesized Multimodal Data).}
We adopt a three-step easy-to-hard sequence built on caption and VQA supervision, then preference optimization:

\textbf{(2a) Shallow Alignment.}  
We start with 1M image–caption pairs, keeping both the vision encoder and LLM component fully frozen. Only the connector and adapter layers are trained. Captioning tasks help establish a stable semantic bridge between vision and language modalities.

\textbf{(2b) Deep Alignment (From Coarse to Full Training).}  
Next, we train on 2M image–QA samples (two questions per image for cross-validation) and 50K GPT-4o-polished samples for supervised fine-tuning, covering recognition, attributes, diagnosis, and basic multi-hop reasoning. Using LoRA, we gradually unfreeze the vision encoder and LLM, enabling transition to full multimodal reasoning.

\textbf{(2c) GRPO Optimization.}  
We build 15K GRPO samples for reinforcement learning. GRPO rewards image-text consistency, internal logic, and verifiable terminology. Details are in Appendix~A.2 and A.6.

\subsection{AgriBench-VL-4K}

\noindent\textbf{Construction.} To ensure the objectivity and reliability of evaluation, we construct \textsc{AgriBench-VL-4K} with an emphasis on data independence, procedural transparency, and rigorous quality control. The benchmark consists of two components: 2{,}018 open-ended question--answer pairs and 1{,}858 single-choice questions. All items are regenerated from held-out images that are never used during training, preventing any overlap with the distributional patterns of the training data generator. Open-ended questions are deeply rewritten from structured captions, covering recognition, symptom/mechanism analysis, management recommendations, and simple multi-step reasoning; answers are normalized for terminology, synonyms, and measurement units. For the single-choice portion, two complementary questions are designed per image, and cross-question consistency is used to reduce random guessing. Distractors are mined from confusable taxa and frequently co-occurring agricultural conditions to increase discriminability and reduce generator-style bias.

\noindent\textbf{Quality control and de-duplication.} To further mitigate potential data contamination, we perform strict de-duplication at both the image and text levels: perceptual hashing and visual-feature similarity are used to remove near-duplicate images, and lexical as well as embedding similarity are applied to eliminate near-duplicate question--answer strings. De-duplication is conducted both across train--evaluation splits and within the evaluation split itself, ensuring that \textsc{AgriBench-VL-4K} contains no residual patterns originating from the training set or the data generator. All remaining items then undergo manual review by annotators with backgrounds in agriculture and computer science, who jointly verify factual correctness, image-grounded evidence, and ambiguity resolution. In practice, approximately 7.8\% of initial items (4\% for open-ended QA and 11.5\% for single-choice QA) are rejected due to factual inconsistencies or lack of image relevance.

\section{Results}

\subsection{Comparative Experiment}

\begin{table*}[t]
\centering
\small
\caption{Language-only evaluation (text capability only). Comparison of AgriGPT-VL with general VLMs on text tasks without images. \textbf{Bold} and \underline{underlined} denote best and second-best per column.}

\label{tab:Text-only evaluation}
\begin{tabular*}{\textwidth}{@{\extracolsep{\fill}}lccccc}
\toprule
\textbf{Model} & \textbf{BLEU} & \textbf{Meteor} & \textbf{Rouge-1-f} & \textbf{Rouge-2-f} & \textbf{Rouge-L-f} \\
\midrule
InternVL-3-8B        & 5.52 & 23.07 & 24.14 & 5.69 & 23.08 \\
InternVL-3-14B       & \underline{8.53} & 27.56 & \underline{26.75} & \textbf{6.46} & \underline{25.56} \\
LLaVA-1.5-7B         & 1.44 & 13.62 & 21.67 & 4.88 & 20.60 \\
MiniCPM-V-2.6-8B        & 1.15 & 12.50 & 21.41 & 5.25 & 20.16 \\
Yi-VL-6B             & 1.03 & 12.38 & 20.93 & 4.36 & 20.09 \\
Yi-VL-34B            & 1.69 & 14.27 & 21.82 & 4.74 & 20.27 \\
Qwen-VL-7B           & 7.70 & 30.17 & 24.16 & 4.97 & 22.86 \\
Qwen2.5-VL-72B-Instruct             & 6.52 & \underline{30.27} & 25.84 & 5.95 & 24.43 \\
Qwen/QVQ-72B-Preview            & 2.48 & 17.31 & 17.54 & 3.63 & 16.75 \\
DeepSeek-VL-1.2      & 6.37 & 29.67 & 22.10 & 4.93 & 20.93 \\
Gemini-2.5-Flash     & 6.12 & 27.49 & 24.85 & 5.59 & 23.73 \\
Gemini-2.5-Pro       & 4.34 & 23.69 & 24.16 & 4.45 & 22.30 \\
AgriGPT-VL (ours)    & \textbf{10.84} & \textbf{32.53} & \textbf{27.73} & \underline{6.36} & \textbf{26.36} \\
\bottomrule
\end{tabular*}
\end{table*}

\begin{table*}[t]
\centering
\small

\caption{Performance comparison on the AgriBench-VL-4K benchmark. 
\textbf{Acc} and \textbf{Acc$^{+}$} correspond to accuracy on single-choice visual reasoning questions, whereas the other metrics evaluate open-ended question–answering quality.}

\label{tab:VL-eval}

\begin{tabular*}{\textwidth}{@{\extracolsep{\fill}}lccccccc}
\toprule
\textbf{Model} & \textbf{Acc} & \textbf{Acc$^+$} & \textbf{BLEU} & \textbf{Meteor} & \textbf{Rouge-1-f} & \textbf{Rouge-2-f} & \textbf{Rouge-L-f} \\
\midrule
InternVL-3-8B        & 81.27\% & 66.31\% & 5.38 & 21.85 & 33.43 & 12.22 & 30.69 \\
InternVL-3-14B       & \underline{83.05\%} & \underline{69.21\%} & 6.32 & 23.26 & 35.01 & \underline{13.44} & 32.11 \\
LLaVA-1.5-7B         & 62.33\% & 40.04\% & 2.39 & 15.55 & 30.96 & 11.09 & 28.57 \\
MiniCPM-V-2.6-8B        & 76.53\% & 59.63\% & 7.12 & 22.57 & 36.02 & 13.39 & 33.36 \\
Yi-VL-6B             & 63.89\% & 40.69\% & 2.21 & 15.21 & 30.32 & 10.59 & 27.88 \\
Yi-VL-34B            & 69.48\% & 48.98\% & 2.83 & 16.61 & 32.05 & 10.98 & 29.42 \\
Qwen2.5-VL-72B-Instruct & 81.70\% & 67.49\% & \underline{15.41} & \underline{41.38} & \underline{39.14} & 13.25 & \underline{36.63} \\
Qwen/QVQ-72B-Preview & 76.00\% & 58.99\% & 4.55 & 19.43 & 31.46 & 9.09 & 29.53 \\
DeepSeek-VL2         & 79.49\% & 63.72\% & 2.88 & 30.86 & 25.57 & 7.21 & 22.97 \\
Gemini-2.5-Flash     & 80.68\% & 65.88\% & 9.12 & 29.38 & 33.47 & 10.00 & 31.33 \\
Gemini-2.5-Pro       & 81.59\% & 66.74\% & 6.55 & 26.22 & 30.25 & 8.65 & 28.01 \\
\textbf{AgriGPT-VL (ours)} & \textbf{85.84\%} & \textbf{74.17\%} & \textbf{26.27} & \textbf{47.55} & \textbf{46.52} & \textbf{20.09} & \textbf{43.81} \\
\bottomrule
\end{tabular*}

\end{table*}

We focus on two questions: (i) after progressively injecting domain knowledge and vision--language alignment, is textual competence preserved and strengthened; and (ii) in real image--language settings, does the model exhibit stronger visual grounding and agronomic reasoning---i.e., can it both choose correctly (discriminative robustness) and articulate evidence--based answers (generation quality).
To this end, we evaluate text--only capability on AgriBench-13K \citep{Yang2025AgriGPT}and multimodal capability on AgriBench-VL-4K.
For discriminative evaluation, we report \emph{Acc} (single--choice accuracy, scored per question) and \emph{Acc$^+$} (image--level cross--consistency: both single--choice questions for the same image must be correct).
For generation, we report BLEU \citep{papineni2002bleu}, METEOR \citep{banerjee2005meteor}, and ROUGE-L \citep{lin2004rouge} to measure terminology conformity, semantic coverage, and structural completeness.

 We compare \textit{AgriGPT-VL} against twelve representative vision--language models:
InternVL-3-8B/3-14B \citep{zhu2025internvl3}, 
LLaVA-1.5-7B \citep{liu2023llava15},
MiniCPM-V-2.6 \citep{yao2024minicpmv,openbmb2024minicpmv26},
Yi-VL-6B and Yi-VL-34B \citep{yi2024yivl6b,yi2024yivl34b},
Qwen-VL-7B \citep{bai2023qwenvl},
Qwen2.5-VL-72B-Instruct \citep{bai2025qwen25vl},
Qwen-QVQ \citep{qwen2024qvq},
DeepSeek-VL-1.2 \citep{lu2024deepseekvl},
and Gemini-2.5 Flash/Pro \citep{google2025gemini25flash,google2025gemini25pro}.

As shown in tables~\ref{tab:Text-only evaluation}, on AgriBench-13K  \citep{Yang2025AgriGPT}, AgriGPT-VL leads across mainstream text metrics, indicating that the progressive training does not sacrifice language ability; instead, it strengthens standardized use of agricultural terminology and canonical answer style, consolidating textual representations and providing a stable linguistic base for the subsequent multimodal stage.

As shown in table~\ref{tab:VL-eval}, on \textsc{AgriBench-VL-4K}, we obtain the best results on all metrics, surpassing several flagship large models. Gains in \emph{Acc} reflect more precise image--option matching; gains in \emph{Acc$^+$} demonstrate consistent semantics per image and stronger resistance to hard distractors (confusable taxa and co--occurring conditions), thereby mitigating chance guessing and better reflecting true capability. Improvements in Bleu \citep{papineni2002bleu}, Meteor \citep{banerjee2005meteor}, and Rouge-1-f/Rouge-2-f/Rouge-L-f \citep{lin2004rouge} further indicate three strengthened abilities: (1) visual evidence grounding and factor extraction (organs, colors/lesions, phenology); (2) agronomic multi-step reasoning (from symptoms to plausible causes and management consistent with scene constraints); and (3) professional, audit-ready expression (units, terminology, and thresholds that follow domain conventions).Detailed definitions and computation formulas of the evaluation metrics are included in Appendix A.5

In addition, as shown in figure~\ref{fig:LLM based Judger}, we conduct JudgeLM  \citep{jiao2023panda}  blind pairwise comparisons: for each query, two systems’ outputs are judged head-to-head, we swap left/right positions to reduce order bias, and average the two outcomes.
We report three preference metrics: \emph{WR} (ties excluded).
Across most strong baselines, AgriGPT-VL achieves consistently higher win rates and remains competitive against top large models, corroborating the above advantages from a preference perspective. Appendix A.3 describes the prompt design for the LLM-based judger, and Appendix A.4 details the metric computation methodology.

\subsection{Ablation Study}

\begin{table*}[t]
\centering
\small
\setlength{\tabcolsep}{4.5pt}
\caption{Ablation study of alignment stages. \textbf{Bold} indicates best per column.}
\label{tab:ablation_evluation}

\begin{tabular*}{\textwidth}{@{\extracolsep{\fill}}lccccccc}
\toprule
\textbf{Setting} & \textbf{Acc} & \textbf{Acc$^+$} & \textbf{BLEU} & \textbf{Meteor} & \textbf{Rouge-1-f} & \textbf{Rouge-2-f} & \textbf{Rouge-L-f} \\
\midrule
Base (Qwen2.5-VL-7B)            & 77.20\% & 60.32\% & 13.42 & 38.24 & 35.52 & 10.78 & 32.73 \\
+ Shallow Alignment             & 78.23\% & 62.47\% & 15.54 & 36.47 & 40.76 & 14.07 & 37.95 \\
+ Shallow + Deep                & 81.18\% & 66.67\% & 21.68 & 44.38 & 43.04 & 15.62 & 40.37 \\
+ Shallow + Deep + GRPO         & \textbf{85.84\%} & \textbf{74.17\%} & \textbf{26.27} & \textbf{47.55} & \textbf{46.52} & \textbf{20.09} & \textbf{43.81} \\
\bottomrule
\end{tabular*}

\end{table*}

\begin{table*}[t]
\centering
\small
\setlength{\tabcolsep}{4.5pt}
\caption{Evaluation of general capabilities before and after fine-tuning.}
\label{tab:generalization_test}

\begin{tabular*}{\textwidth}{@{\extracolsep{\fill}}lcccccc}
\toprule
\textbf{Model} & \textbf{MMLU} & \textbf{ARC} & \textbf{OpenBookQA} & \textbf{MMBench} & \textbf{MMMU} & \textbf{SeedBench} \\

\midrule
Qwen2.5-VL-7B & 0.6783 & 0.9043 & 0.8501 & 0.8398 & 0.4329 & 0.7565 \\
AgriGPT-VL(7B) & 0.6741 & 0.8462 & 0.8412 & 0.8312 & 0.4599 & 0.7574 \\
\bottomrule
\end{tabular*}

\end{table*}

Starting from a base model, we progressively add \emph{Shallow Alignment} (caption-only supervision with the vision stack frozen to establish cross-modal semantic anchors), \emph{Deep Alignment} (single-choice reasoning with the vision encoder and cross-modal interaction layers unfrozen), and \emph{GRPO} (reinforcement optimization with 15k GRPO reinforcement learning dataset). 

As shown in tables~\ref{tab:ablation_evluation}, the results reveal a clear hierarchy of contributions: Shallow Alignment primarily improves lexical and descriptive consistency, stabilizing image--text keypoint correspondence; Deep Alignment is the main driver of cross-modal understanding and reasoning, lifting both discriminative and generation metrics; and GRPO further enhances factual faithfulness and robustness, with the largest gains on the stricter image-level cross-consistency metric (\emph{Acc$^+$}), indicating that expert-level instructions are necessary to constrain high-precision behavior.

\subsection{Generalization Evaluation}

To assess whether domain specialization preserves general capabilities, we compare the fine-tuned \textit{AgriGPT-VL} with its base model (\textit{Qwen2.5-VL}) on six public benchmarks: three text-only (\textsc{MMLU}~\citep{hendrycks2021mmlu}, \textsc{ARC}~\citep{clark2018arc}, \textsc{OpenBookQA}~\citep{mihaylov2018can}) and three vision--language (\textsc{MMBench}~\citep{liu2025mmbench}, \textsc{MMMU}~\citep{Yue_2024_CVPR}, \textsc{SEED-Bench}~\citep{Li_2024_CVPR}).

Overall, \textit{AgriGPT-VL} remains competitive. On text-only tasks, performance is largely preserved on \textsc{MMLU} and \textsc{OpenBookQA}, with only a modest decline on \textsc{ARC}. On vision--language tasks, the model matches or exceeds the base, showing parity on \textsc{SEED-Bench} and \textsc{MMBench}, and clear gains on \textsc{MMMU}.

As shown in Table~\ref{tab:generalization_test}, two conclusions emerge: (i) the curriculum—starting with textual grounding—effectively mitigates forgetting, maintaining broad competence across $\sim$40K out-of-domain samples; (ii) the improvement on \textsc{MMMU} confirms that learned visual reasoning generalizes beyond agriculture, reinforcing the strength and transferability of our finetuning framework.

\subsection{External Evaluation on AgriBench}

To further assess the robustness of our model, we conduct an additional evaluation on AgriBench\citep{zhou2024agribench}, an external benchmark comprising approximately 700 single-choice questions. Unlike our in-house \textsc{AgriBench-VL-4K}, AgriBench differs entirely in data sources, annotation protocols, textual style, and question distribution, and is therefore fully decoupled from our training data and data-generation pipeline. This provides a genuinely out-of-distribution setting for examining whether the model can transfer across construction paradigms. Such an external evaluation not only eliminates potential biases arising from stylistic alignment but also offers a more comprehensive measure of the model’s ability to interpret real agricultural scenarios.

As shown in tables~\ref{tab:agribench_acc}, AgriGPT-VL achieves the highest accuracy of (70.10\%) on AgriBench\citep{zhou2024agribench}, clearly outperforming publicly available vision--language models including Gemini-2.5-Pro (56.94\%). The significance of this independent evaluation lies in demonstrating that the improvements of AgriGPT-VL do not rely on the stylistic patterns of our training data nor on overfitting to a single construction pipeline. Instead, the model exhibits stronger grounding in agricultural knowledge, pest and disease characteristics, and field-level visual cues.
For agriculture, achieving clear gains on a fully external benchmark indicates robust cross-source generalization.

\begin{table}[t]
\centering
\small
\setlength{\tabcolsep}{3pt} 
\caption{Model accuracy on AgriBench dataset.}
\label{tab:agribench_acc}

\begin{tabular*}{\columnwidth}{@{\extracolsep{\fill}} l r }
\toprule
\textbf{Model} & \textbf{Acc} \\
\midrule
InternVL-3-8B               & 65.66\% \\
InternVL-3-14B              & \underline{60.08\%} \\
LLaVA-1.5-7B                & 59.29\% \\
Qwen2.5-VL-7B               & 49.94\% \\
Qwen2.5-VL-72B-Instruct     & 55.64\% \\
Qwen/QVQ-72B-Preview        & 45.39\% \\
DeepSeek-VL2                & 50.84\% \\
Gemini-2.5-Flash            & 53.70\% \\
Gemini-2.5-Pro              & 56.94\% \\
\textbf{AgriGPT-VL (ours)}  & \textbf{70.10\%} \\
\bottomrule
\end{tabular*}
\vspace{-8pt}
\end{table}

\section{Conclusion}

We present AgriGPT-VL, an agricultural vision–language understanding suite that unifies large-scale data generation, curriculum-based multimodal training, and benchmark evaluation. The model demonstrates strong agronomic reasoning and visual grounding without sacrificing general capabilities. This compact and reproducible framework provides a practical blueprint for building specialized multimodal systems in agriculture and beyond.

{
    \small
    \bibliographystyle{ieeenat_fullname}
    \bibliography{main}
}

% WARNING: do not forget to delete the supplementary pages from your submission 
% \input{sec/X_suppl}

\end{document}